\documentclass{article} 

\usepackage[a4paper, margin=1in]{geometry}
\usepackage{graphicx}
\usepackage{pdflscape}
\usepackage{subcaption}

\usepackage{amsmath}
\usepackage{amssymb}
\usepackage{amsthm}

\usepackage{tikz}
\usetikzlibrary{positioning, arrows.meta, calc, fit}

\usepackage{hyperref}
\usepackage{url}

\usepackage{algorithm}
\usepackage[]{algpseudocode}

\usepackage{authblk}

\title{ExMAG: Learning of Maximally Ancestral Graphs}

\author[1]{Petr Ryšavý}
\author[1]{Pavel Rytíř}
\author[1]{Xiaoyu He}
\author[2,1,3]{Georgios Korpas}
\author[1]{Jakub Mareček}

\affil[1]{Department of Computer Science,
Czech Technical University in Prague, Czech Republic}
\affil[2]{HSBC Quantum Technologies Group, Innovation \& Ventures, HSBC, Singapore}
\affil[3]{Archimedes Research Unit on AI, Data Science and Algorithms, Marousi, Greece}

\begin{document}

\maketitle

\begin{abstract}
In mixed graphs, there are both directed and bidirected edges. An extension of acyclicity to this mixed-graph setting is known as maximally ancestral graphs. 
This extension is of considerable interest in causal learning in the presence of confounders. There, directed edges represent a clear direction of causality, while bidirected edges represent confounding.

We propose a branch-and-cut algorithm for learning maximally ancestral graphs using a formulation as a mixed-integer quadratic program. Empirically, our method achieves comparable or improved reconstruction quality while requiring an order of magnitude fewer samples than state-of-the-art approaches.
\end{abstract}

\section{Introduction}
\label{section:introduction}

As one transitions from statistical to causal learning \cite{scholkopf2022statistical}, one is seeking the most appropriate causal model. Bayesian networks (BN)  \cite{pearl1985bayesian} are a popular model, where a directed acyclic graph represents causal relationships. The vertices represent random variables, and the absence of oriented edges encodes conditional variable independence and, therefore, the edges suggest the causal relationships. The conditional probability distributions assigned to vertices quantify the strength of the causal relationships. While the learned graph needs to be acyclic, a major challenge in learning BNs is confounding.

Simpson's paradox shows that without considering confounding factors in statistical analysis \cite{McElreath2018Statistical}, the direction of causality can be misestimated completely. An example of a textbook \cite{McElreath2018Statistical} comes from the Berkeley graduate admissions \cite{bickel1977sex}. The data show that women find it harder to get admitted to Berkeley graduate schools. Nevertheless, this is because women tend to apply to departments that have lower admission rates. 
In this example, the choice of the graduate school is the confounder, impacting the probability of admission.
Confounding is prevalent throughout high-dimensional statistics \cite{Lin2014Comparison,Gilad2015reanalysis}.



Specifically, in biomedical sciences, confounders such as socio-economic status, age, or lifestyle factors can distort the true causal relationship between treatments and outcomes \cite{zhou2022causal}. Techniques such as instrumental variables \cite{reiersol1945confluence,Imbens2014Instrumental}, propensity score matching \cite{rosenbaum1983central}, and double machine learning \cite{chernozhukov2018double} have been widely used to mitigate the effects of confounding in clinical trials and observational studies.
To mitigate confounding biases, statistical models that explicitly account for hidden confounders, such as spectral methods and latent variable models, are often employed \cite{Guo2022doubly}. Furthermore, meta-analysis and sensitivity analysis are often used to evaluate the robustness of findings in the presence of potential confounders, especially when combining results from multiple studies \cite{Buhlmann2020Invariance, Mathur2022Methods}. 


Confounding arises when an unobserved or uncontrolled variable influences multiple observed variables simultaneously, thereby creating associations that need not correspond to direct causal effects \cite{Buhlmann2011Statistics,zhou2022causal}. In causal structure learning, this is particularly challenging because hidden confounders can both obscure genuine causal directions and induce spurious dependencies among observed variables. Classical strategies such as instrumental variables, propensity score methods, double machine learning, and latent-variable adjustments can reduce the statistical impact of confounding \cite{reiersol1945confluence,Imbens2014Instrumental,rosenbaum1983central,chernozhukov2018double,Guo2022doubly,Buhlmann2020deconfounding}, but they do not by themselves produce a graphical representation of the underlying causal system. This motivates learning graph classes that encode latent confounding explicitly, especially in partially observed settings where assumptions such as full observability are unrealistic \cite{Wang2021Causal,bhattacharya2021differentiable}.


Maximal Ancestral Graphs (MAGs) \cite{Richardson2002Ancestral} provide a natural extension of Bayesian networks to settings with latent confounding. Rather than estimating a Directed Acyclic Graph (DAG) alone, one can estimate a MAG, which represents both direct causal effects and dependencies induced by hidden confounders through directed and bidirected edges. This richer representation allows MAGs to capture relationships that DAG-based models cannot express without explicitly introducing latent variables, making them better suited to partially observed causal systems.

There are only a few studies of MAG estimation \cite{chen2021integer,Rantanen2021exactsearch,Claassen2022Search,hu2023causal,hu2024towards,hu2024fast,dash2025integer}.
Methods in \cite{Richardson2014Factorization,ommen2024efficiently} are applicable to both discrete and nonparametric cases, which extend DAG to MAG or ADMG diagrams. 
Factorization in MAG is not directly decomposable into individual variables and their parent sets, as in DAGs, but must instead consider components connected by bidirected paths (termed \textit{districts} or \textit{c-components}), cf. \cite{Richardson2014Factorization}, although  \cite{Claassen2022Search} proposed to use Markov equivalence classes (MEC) instead.
In 2021, \cite{chen2021integer} introduced a first mixed-integer programming (MIP) formulation, but the number of variables scales with the number of c-components, i.~e., exponentially with the number of vertices in the worst case. Such formulations are also known as extended formulations \cite{conforti2010extended}. The same year, \cite{Rantanen2021exactsearch} explored a score-based approach for directed MAG discovery, leveraging a local score function optimized using pruning rules and dynamic programming. Additionally, \cite{zhou2022causal} addresses exogenous covariates in causal formulation that helps explain the heterogeneity in both sampling and causal mechanisms. \cite{hu2023causal}'s disseration presented an extension of the imsets of \cite{studeny2006probabilistic} from directed acyclic graphs (DAGs) to towards MAGs \cite{hu2024towards}, which allows for the use of the methods of Studen{\' y}, and a score-based heuristic
\cite{hu2024fast}. 
More recently, \cite{dash2025integer} enhanced the scalability of methods of \cite{chen2021integer} by utilizing linear programming (LP) relaxations instead of solving the MIP.

Our approach proposes a formulation of MAG estimation within Mixed-Integer Quadratic Programming (MIQP). From the exponential set of constraints enforcing acyclicity, only those violated by a solution are iteratively added to the program in a lazy manner. As a result, the initial program is polynomial in the number of vertices, while the worst-case scenario with exponentially many constraints is avoided in many instances of the program. In contrast, the so-called extended formulations of \cite{chen2021integer,dash2025integer} have the dimension exponential in the number of vertices. While both the extended formulation of \cite{chen2021integer} and ours ensure that confounding factors are properly accounted for and the true underlying data are better represented by the model, our implementation scales further, from 4-5 variables in the extended formulation of \cite{chen2021integer} to 25 or more variables with the proposed compact formulation.


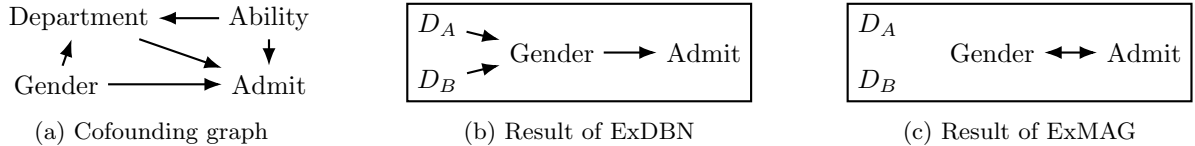
\begin{figure}[tb]
    \centering
    \begin{subfigure}[b]{0.23\textwidth}
         \centering
         \begin{tikzpicture}[scale=0.1]
        
            \node (U) {Ability};
            \node (D) [left=0.8cm of U] {Department};
            \node (A) [below=0.35cm of U] {Admit};
            \node (G) [left=1.5 cm of A] {Gender};
        
            \draw[->, thick, >=Latex] (U) -- (D);
            \draw[->, thick, >=Latex] (U) -- (A);
            \draw[->, thick, >=Latex] (D) -- (A);
            \draw[->, thick, >=Latex] (G) -- (D);
        
        \end{tikzpicture}
         \caption{Confounding graph}
         \label{fig:1a}
     \end{subfigure}
     \hfill    
    \newcommand{\vdst}{0.2}
    \newcommand{\hdst}{0.7}
    \begin{subfigure}[b]{0.36\textwidth}
    \centering
    \begin{tikzpicture}
        \node (DA) {$D_A$};
        \node (DB) [below=\vdst of DA] {$D_B$};
        \path let \p1 = ($(DA)!0.5!(DB)$) in node (Midpoint) at (\x1,\y1) {};
        \node (gen) [right=\hdst of Midpoint] {Gender};
        \node (adm) [right=\hdst of gen] {Admit};
        \draw[->, thick, >=Latex] (DA) -- (gen);
        \draw[->, thick, >=Latex] (DB) -- (gen);
        \draw[->, thick, >=Latex] (gen) -- (adm);
        \node[draw=black, thick, fit=(DA)(DB)(gen)(adm), inner sep=0.005cm] {};
    \end{tikzpicture}
    \caption{Result of ExDBN \cite{exdbn}}
         \label{fig:1b}
     \end{subfigure}
     \hfill
    \begin{subfigure}[b]{0.3\textwidth}
    \centering
    \begin{tikzpicture}
        \node (DA) {$D_A$};
        \node (DB) [below=\vdst of DA] {$D_B$};
        \path let \p1 = ($(DA)!0.5!(DB)$) in node (Midpoint) at (\x1,\y1) {};
        \node (gen) [right=\hdst of Midpoint] {Gender};
        \node (adm) [right=\hdst of gen] {Admit};
        \draw[<->, thick, >=Latex] (gen) -- (adm);
        \node[draw=black, thick, fit=(DA)(DB)(gen)(adm), inner sep=0.005cm] {};
    \end{tikzpicture}
    \caption{Result of ExMAG}
         \label{fig:1c}
     \end{subfigure}
     \hfill
    \caption{Ground truth with the confounder of Department on the Berkeley graduate admission example (left, \ref{fig:1a}), a dynamic Bayesian network trained on the data (center, \ref{fig:1b}), and ExMAG output  (right, \ref{fig:1c}). While the Bayesian network suggests a causal relationship between gender and admission, ExMAG identifies only a cofounder between gender and admission. See the supplementary material for details. }
    \label{fig:Berkeley}
\end{figure}

\subsection{Motivating Example}

Let us revisit the Berkeley graduate admission paradox example. As in most paradoxes, there is no violation of logic in Simpson's paradox, just a violation of intuition. 
In this case, the intuition is that a positive association in the entire population should also hold within each department. 
Overall, females in these data did have a harder time getting admitted to graduate school. But that arose, because female applicants chose the departments that were the most difficult to gain admission to for anyone, male or female. In this example, gender influences the choice of department, and the department influences the chance of admission. Controlling for department reveals a more plausible direct causal influence of gender, as illustrated in Fig. \ref{fig:1a}.
Our method, ExMAG, is able to reveal the confounders in this Berkeley graduate admission example, as illustrated in the notebook in the supplementary materials and Fig.~\ref{fig:Berkeley}.



\section{Graphs and Properties}
\label{section:graphs}
A Directed Acyclic Graph (DAG) is a directed graph \( \mathcal{G} = (\mathcal{V}, \mathcal{E}) \) such that there are no directed cycles. That is, there is no sequence of distinct vertices \( v_1, v_2, \dots, v_k \in \mathcal{V} \) such that \( (v_i, v_{i+1}) \in \mathcal{E} \) for all \( 1 \leq i \leq k-1 \) and \( (v_k, v_1) \in \mathcal{E} \). Maximal ancestral graphs (MAGs), introduced by \cite{Richardson2002Ancestral}, provide a framework for modeling distributions through conditional independence relations. Compared with DAGs, MAGs allow for latent confounders, accommodating data that arise from distributions with more complex independence structures and revealing hidden states in the graphs. 
While DAGs allow for the efficient computation of maximum likelihood estimates (MLEs) and scoring (e.g., via BIC), these properties are challenging to extend to MAG due to their structural and computational complexity \cite{Hu2020Markov}.


\paragraph{ADMG} Mixed graphs feature two types of edges: directed (\( \rightarrow \)) and bidirected (\( \leftrightarrow \)). Mixed graph \( \mathcal{G} \)  thus consists of vertex set \( \mathcal{V} \), a set of directed edges $\mathcal{E}$ and bidirected
edges $\mathcal{B}$, where $\mathcal{E}$ are ordered pairs of vertices,  while  $\mathcal{B}$ are 
unordered 2-element subsets of distinct vertices. 
For vertex \( v \) in $\mathcal{V}$, we define the \textit{parents}, \textit{spouses}, and \textit{ancestors} 
of \( v \), respectively as:
\begin{align*}
\text{pa}_\mathcal{G}(v) &= \{w : w \to v \text{ in } \mathcal{G} \}, \\
\text{an}_\mathcal{G}(v) &= \{w : w \to \cdots \to v \text{ in } \mathcal{G} \text{ or } w = v \}, \\
\text{sp}_\mathcal{G}(v) &= \{ w : w \leftrightarrow v \text{ in } \mathcal{G} \}. &
\end{align*}
As in a DAG, a mixed graph is acyclic if it contains no directed cycles in $\mathcal{E}$, i.e., an acyclic directed mixed graph (ADMG) \cite{hu2023causal}.

A directed mixed graph \(\mathcal{G}\) is ancestral ADMG if it has no directed cycles and no vertex is an ancestor of any of its spouses. Formally,
\[
\text{if } v \neq w \text{ and } v \in \text{an}_\mathcal{G}(w) \cup \text{sp}_\mathcal{G}(w), \text{ then } w \notin \text{an}_\mathcal{G}(v).
\] Equivalently, \(\mathcal{G}\) is an ancestral ADMG if it contains no directed cycles or almost directed cycles \cite{chen2021integer, hu2023causal}. 
In an ADMG, an almost directed cycle is of the form \(v \to u \to \ldots \to w \leftrightarrow v\); in other words, \(\{w, v\} \in \mathcal{B}\) is a bidirected edge, and \(v \in \text{an}_\mathcal{G}(w)\) \cite{chen2021integer}.


A vertex $u$ on a non-overlapping path is called a collider if it is contained in an non-overlapping subpath $(w,u,v)$ with two arrowheads into $u$. In mathematical form, $u$ is collider on path $P = p_1p_2\cdots w u v \cdots p_{k}$ if and only if either $w\rightarrow u$ or $w \leftrightarrow u$, and at the same time $u \leftarrow v$ or $u \leftrightarrow w$.

\paragraph{Inducing Paths}
A path $P=(v_1,\ldots,v_{k})$ in an ADMG is an inducing path~\cite{richardsonAncestralGraphMarkov2002a,zhangCausalReasoningAncestral} if every non-endpoint vertex on $P$ is a collider and an ancestor of at least one of $v_1$ or~$v_{k}$.

\paragraph{m-separation}
Graphs encode conditional independence via separation criteria. For acyclic directed mixed graphs (ADMGs), \textit{m-separation} generalizes d-separation to handle bidirected edges. A path between vertices \( u \) and \( v \) is \textit{m-connecting} given a conditioning set \( C \subseteq \mathcal{V} \) if: (i) \( u \) and \( v \) are the endpoints; (ii) all non-colliders are not in \( C \); and (iii) all colliders are in \( \operatorname{an}_\mathcal{G}(C) \). Vertices \( u \) and \( v \) are \textit{m-separated} given \( C \) if no such path exists.


\paragraph{ Maximal Ancestral Graph} An \emph{ancestral} ADMG $\mathcal{G}$ is called a maximal ancestral graph (MAG) if the \emph{maximality} condition holds, i.e., for every pair of nonadjacent vertices $u$ and $v$, there exists some set $C$ such that $u, v$ are \textit{m-separated} given $C$ in $\mathcal{G}$. We refer to \cite{hu2023causal} for multiple examples. The existence of a directed cycle or an almost directed cycle contradicts ancestrality. Moreover, in an ancestral ADMG, the existence of an inducing path between two non-adjacent vertices implies that the graph is not maximal.


\section{Structural Equation Model with Latent Confounders}

Graphical causal models provide a principled framework for representing causal relationships and reasoning about conditional independence structures among observed variables in the presence of latent variables. Here, we adopt the Structural Equation Model (SEM) framework \cite{Bollen1989Structural, Pearl2009Causality}, which allows causal relationships among variables to be represented through structural equations while explicitly accommodating unobserved factors. Under this framework, the data are generated so that

$$
X \leftarrow X W_D + LW_L + \epsilon,
$$


where
$X=(X_1,\dots,X_d)$ is the vector of variables, $L=(L_1,\dots,L_p)$ is the vector of latent variables, $W_D$ is a $\mathbb{R}^{d\times d}$ weighted adjacency matrix of the DAG encoding the causal structure, $W_L\in\mathbb{R}^{p\times d}$ encodes structure of latent variables, and $\epsilon$ is the noise vector.



Further, we will assume that the latent confounding term $LW_L$ has been absorbed into the error term~$\epsilon$.
$$
X \leftarrow X W_D + \epsilon.
$$
The covariance matrix $B=\mathbb{E}[\epsilon \epsilon^{T}]$ will represent the hidden confounders.

If each $\epsilon_i$ is Gaussian, then the resulting distribution
$p(X)$ is multivariate normal with zero mean and covariance
\[
  \Sigma = (I - W_D)^{-T} W_B (I - W_D)^{-1} .
\]



The cost function then becomes the error of the prediction, represented by the difference between the sampled data $X$ and their prediction stemming from observed variables, and influece of latent confounders, similarly to \cite{bhattacharya2021differentiable}. See derivation in the Supplementary materials.
$$
J(W_D,W_B)=\lVert X-XW_D - \left( X-XW_D \right)W_B\rVert^2.
$$

\subsection{Connection to Causality}
\label{subsection:Causality}
The causal parameters $W_D$ and $W_B$ are solution to the following minimization problem:
\[
\arg\min_{W_D, W_B} \max_{P \in \mathcal{P}} \mathbb{E}_P \left[(X - X W_D - \left( X-XW_D \right)W_B)^2\right],
\]
where $\mathcal{P}$ is a class of distributions containing perturbations of the original distribution, including confoundings. Class $\mathcal{P}$, therefore, guides the overall structure of the MIQP formulation. 

Now we denote by $X$ the data matrix. Let $X \in \mathbb{R}^{n \times d}$ be an $n\times d$ matrix with data samples. Then, under the assumption of Gaussian noise, the problem can be reformulated as
\begin{equation}
\arg\min_{W_B, W_D} J(W_D, W_B).
\label{eq:mincostfunct}
\end{equation}

\section{Formulation of the Mixed Integer Quadratic Program}

Here, we present the formulation of the Mixed Integer Quadratic Program (MIQP) used to infer the causal structure, with binary matrix \( B = [b_{j,k}] \in \{0,1\}^{d \times d} \) introduced to account for relationships explained by confounding factors, alongside 
binary adjacency matrix \(E = [e_{j,k}] \in \{0,1\}^{d \times d} \) adopted from the ExDAG  model~\cite{Rytivr2024exdag}. Two weight matrices $W_D = [w_{j,k}] \in \mathbb{R}^{d \times d}$ and $W_B = [w_{j,k}] \in \mathbb{R}^{d \times d}$ encode the weights of directed and bidirected edges. Whenever entry ${w_D}_{j,k}$ (${w_B}_{j,k}$, respectively) is non-zero, $e_{j,k}$ (directed edge) (or $b_{j,k}$ (bidirected edge), respectively) is nonzero. At the same time, we extend the existing formulation by introducing an additional optional binary input matrix \( F = [f_{j,k}] \in \{0,1\}^{d \times d} \), where $f_{j,k} = 1$, indicates that there is no direct causal relationship between variables \(j\) and \(k\), but $j$ and $k$ might have a common cofactor. This follows from the meaning of the edges in a MAG $\rightarrow$ edge implies a direct causal relationship, but does not rule out a possible latent confounding, $\leftrightarrow$ means no direct causal relationship. Such a matrix might be obtained by a statistical test or from background knowledge to improve the results if needed.

Formally, we define \emph{Directed Edge Matrix } $E$ as $e_{j,k} = 1$ if $j \to k$, $0$ otherwise. Similarly, \emph{Bidirected Edge Matrix } $B$ is $b_{j,k} = 1$ if $j \leftrightarrow k$, and $0$ otherwise. Lastly, the input matrix $F$ with pairs of variables which are known not to be in a direct cause-effect relationship, is by definition $f_{j,k} = 1$ if $j \not\to k \wedge k \not\to j$ and $0$ otherwise. Note that this matrix is by definition symmetric.

\subsection{MIQP Formulation}
\label{sec:milp}

The cost function for the Mixed Integer Quadratic Program of ExMAG is the $l_q$ norm below. It has two components - the error of prediction of $X_{i,j}$, and the regularization term. Denote
\begin{equation}
R_{i,j} = X_{i,j} - 
\sum_{k=0; k \neq j}^d
X_{i, k} {w_D}_{k,j}.
\label{eq:errorR}
\end{equation}
Then, the minimization problem in \eqref{eq:mincostfunct} extended by regularization terms corresponds to the minimization of the following cost function
\begin{equation}
\begin{aligned}
\label{obj}
\min_{W_D, W_B, E, B}
\sum_{i=1}^n \sum_{j=1}^d \left|
R_{i,j} - \sum_{k=0; k \neq j}^d R_{i, k} {w_B}_{k,j} 
\right|^q 
+ \lambda \sum_{j=0}^d \sum_{k=0}^d  (e_{j,k} + b_{j,k}),
\end{aligned}
\end{equation}
In this formula, 
\(X_{i,j}\) represents the value of the \(j\)-th variable for the \(i\)-th data point;
\({w_D}_{k,j}\) represents the weight of the directed edge from variable \(k\) to variable \(j\);
\({w_B}_{k,j}\) represents the weight of the bidirected edge from variable \(k\) to variable \(j\);
\(e_{j,k}\) is the binary decision variable indicating the presence of a directed edge from \(j\) to \(k\);
\(b_{j,k}\) is the binary decision variable indicating a bidirected edge between \(j\) and \(k\);
\(\lambda \in \mathbb{R}^+\) is a regularization parameter controlling the model fit and the edge penalty trade-off.
The exponent \( q \in \mathbb{N} \) can take values \( q = 1 \) or \( q = 2 \). 

Optimization criterion in \eqref{obj} implies that the dependencies between the variables are linear. The first part of the criterion encodes for the actual cost as an error of the prediction, the second part encodes for regularization, penalizing more edges with a larger $\lambda$.


As in the ExDAG \cite{Rytivr2024exdag} model, the weights are bounded by introducing large constant \(c\), 
which is chosen large enough to exceed the maximum weight expected in the problem being solved.
The bounding avoids bilinear terms in the cost function in \eqref{obj} and takes the following form:
\begin{align}
\label{constr:Weight}
\tag{Weight Constraint}  
-c \cdot E &\leq W_D \leq c \cdot E,\\
-c \cdot B &\leq W_B \leq c \cdot B,\\
\notag
\label{constr:Edge}
\tag{Edge Constraint}  
E + B &\leq \mathbf{1}.
\notag
\end{align}
The \ref{constr:Edge} means that there cannot be a directed as well as a bidirected edge between the same two vertices. Additionally, we enforce that the bidirected matrix is symmetric by \eqref{eq:symmetry}. If $f_{j, k} = 1$, then we know there is no direct causal relationship between $j$ and $k$, and therefore, $e_{j,k}=0$. This is formally enforced by \eqref{eq:forbidden}. Inversely, $f_{j, k} = 0$ implies a directed edge rather than a bidirected edge between $j$ and $k$ in \eqref{eq:bidirforbi}. Both equations \eqref{eq:forbidden} and \eqref{eq:bidirforbi} are enforced only if the respective fields in matrix $F$ are defined, as they are optional.
\begin{align}
  B &= B^T, \label{eq:symmetry} \\
  F + E &\leq \mathbf{1}, \label{eq:forbidden} \\
  B &\leq F. \label{eq:bidirforbi}
\end{align}
Lastly, we must enforce conditions for directed or almost directed cycles and inducing paths. Those conditions are enforced lazily using a separation routine explained later. Directed cycles are enforced in a way adopted from \cite{Rytivr2024exdag}. Therefore, they are left out of this paper. An almost directed cycle formed by edges in set $E'$ and a bidirected edge $(u,v)$ is forbidden by the constraint
\begin{equation}
\label{constr:Acycliclazy}
\tag{Acyclic Constraint}  
  b_{u, v} + \sum_{(j,k) \in E'} e_{j,k} \leq |E'|.
  \notag
\end{equation}
Similarly, if there is an inducing path formed by path $P$ that contains bidirected edges, and set $E'$ contains all directed edges that participate in the ancestor relationship (including multiple paths) between the inner points of the path and the terminals of $P$, this inducing path is forbidden by
\begin{equation}
\label{constr:InducingPathslazy}
\tag{Inducing-Paths Constraint}  
  \sum_{(j,k) \in P} b_{j,k} + \sum_{(j,k) \in E'} e_{j,k} \leq |E'| + |P| - 1.
  \notag
\end{equation}
Note that the second condition does not necessarily eliminate the inducing path, as the optimizer might forbid one of the edges in $E'$ without influencing the ancestor relationship. This results in path $P$ being found in the next iteration, with a smaller set of directed edges, and the process is repeated.

By enforcing these constraints, we ensure that the MIQP correctly models the causal relationships between the variables while respecting the structure defined by \(e_{j,k}\) and the potential confounding relationships captured by \(b_{j,k}\).

\section{Separation Routine for the Maximal Ancestral Graphs}

The main contribution of this section is the separation routine that identifies whenever a graph is an instance of a maximal ancestral graph. To do so, we need to identify directed cycles, almost directed cycles, and inducing paths. 
The presence of directed cycles can be detected in $\mathcal{O}(d^2)$ using depth-first-search (DFS); such an approach can be found in \cite{Rytivr2024exdag}. For both inducing paths and almost directed cycles, we will use the distance matrix $D$ constructed on the graph of directed edges $E$. This distance matrix can be obtained, for example, using the Floyd-Warshall algorithm \cite{floyd-warshall}.

Having the distance matrix, to check for almost directed cycles, we can iterate over all bidirected edges and test whether the distance between the endpoints using $E$ is finite. If so, we have a directed path connected by a bidirected edge. See Algorithm \ref{alg:almostdirected} for details.

\begin{algorithm}
   \textbf{Input:} directed edges $E$, bidirected edges $B$
   \hrule
  \begin{algorithmic}
    \Function{Almost-Directed-Cycles}{$E$, $B$}
      \State $D \gets $ \Call{Distance-Matrix}{$E$}
      \ForAll{$(j, k) \in \{1,2, \ldots, d\} \times \{1,2, \ldots, d\}$}
          \If{$j \neq k \And b_{j, k} == 1 \And D_{j, k} < \infty$}
             \State $E' = $ \Call{Trace-Distance-Matrix}{$D$,$E$,$j$,$k$}
             \State\Comment{Finds all edges on any $j$ to $k$ path, see Supl.}
             \State{Found cycle formed by edges $E'$ and $j \leftrightarrow k$}
          \EndIf
        \EndFor
    \EndFunction
  \end{algorithmic}
  \caption{Function that identifies almost directed cycles.}
  \label{alg:almostdirected}
\end{algorithm}

\begin{algorithm}
   \textbf{Input:} directed edges $E$, bidirected edges $B$
   \hrule
  \begin{algorithmic}
    \Function{Inducing-Paths}{$E$, $B$}
      \State $D \gets $ \Call{Distance-Matrix}{$E$}
      \ForAll{$s \in {1,2, \ldots, d}$}
        \State\Call{\small Inducing-Paths-DFS}{$D$, $E$, $B$, $s$, $s$, $\{1,2,\ldots d\}$, $[s]$}
      \EndFor
    \EndFunction
    \Function{Inducing-Paths-DFS}{$D$, $E$, $B$, $s$, $u$, possible endpoints, path}
      \If{possible endpoints are empty} \Return \EndIf
      \If{\Call{Len}{path} $> 2 \And u$ in possible endpoints }
         \State \Call{Found-Inducing-Path}{$D$,$E$, path}
         \State \Comment{Edges participating in ancestor relationship are recovered, see Suppl. mat.}
      \EndIf
      \ForAll{$v \in {1,2, \ldots, d}$ such that $e_{u,v}=1$}
        \State {v-endpoints $\gets$ possible endpoints \textbf{if } $D_{v, s} < \infty$ \textbf{ else } possible endpoints $\cap \{ x \mid D_{v,x} < \infty \}$}
        \State {\Call{Inducing-Paths-DFS}{$D$,$E$,$B$,$s$,$v$,v-endpoints,path$+ v$}}
      \EndFor
    \EndFunction
  \end{algorithmic}
  \caption{Simplified function that identifies inducing paths.}
  \label{alg:inducingpaths}
\end{algorithm}

\begin{figure*}[t]
    \centering
    \includegraphics[width=0.6\linewidth]{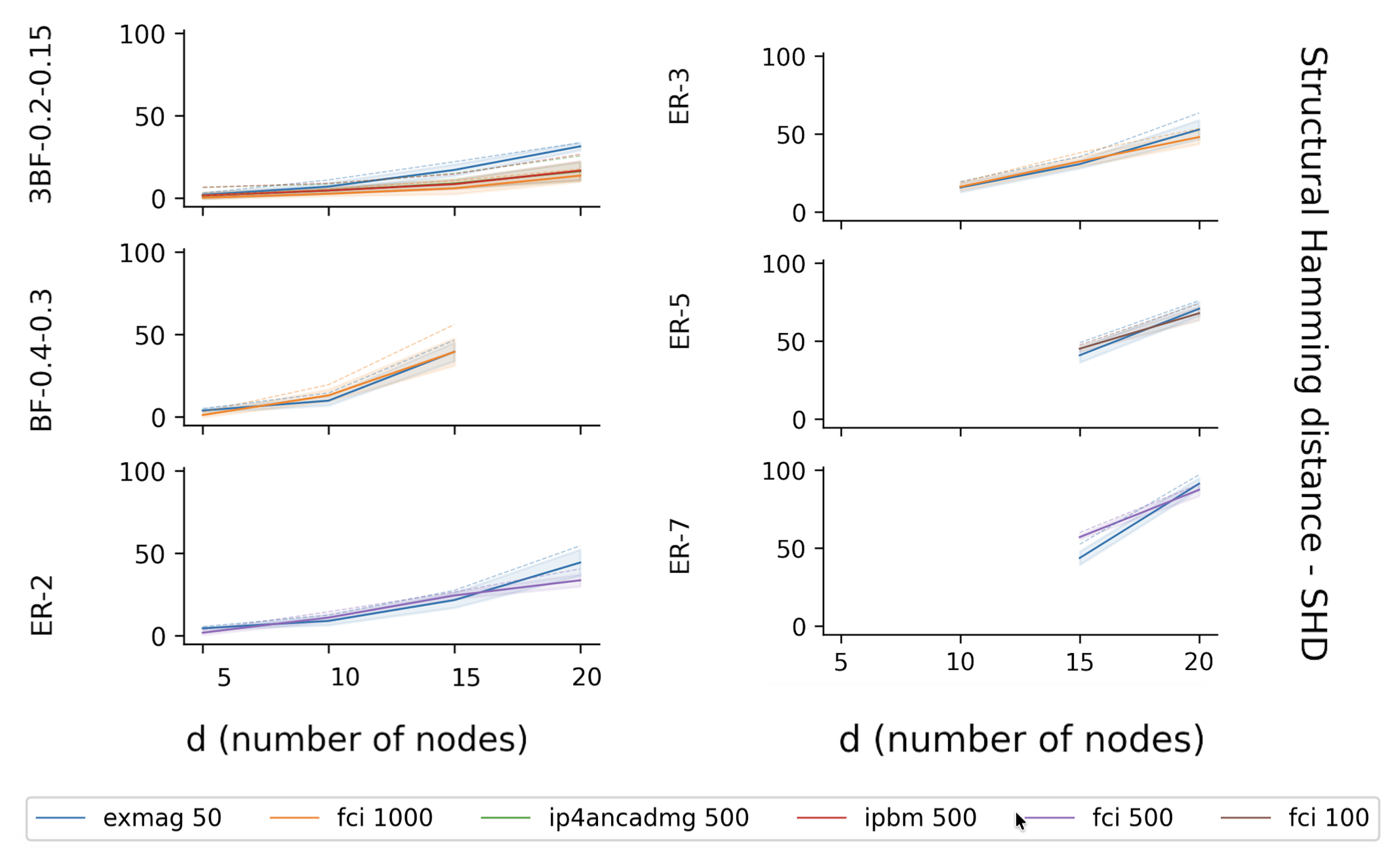}
    \caption{
    SHD values (in the vertical axis) for different settings of $d$ (in the horizontal axis) and different choices of the graph. Standard deviations are depicted as the blurred regions, and dashed lines are the maximum values. For each method, the number provided is the number of samples $n$, and the plot shows primarily the best results over the choice of $n$, illustrating that ExMAG is able to provide competitive results with fewer samples than other methods. See supplementary materials for results on more datasets and error information.
    }
    \label{fig:bstshd}
\end{figure*}
\begin{figure*}[h]
    \centering
    \includegraphics[width=0.8\linewidth]{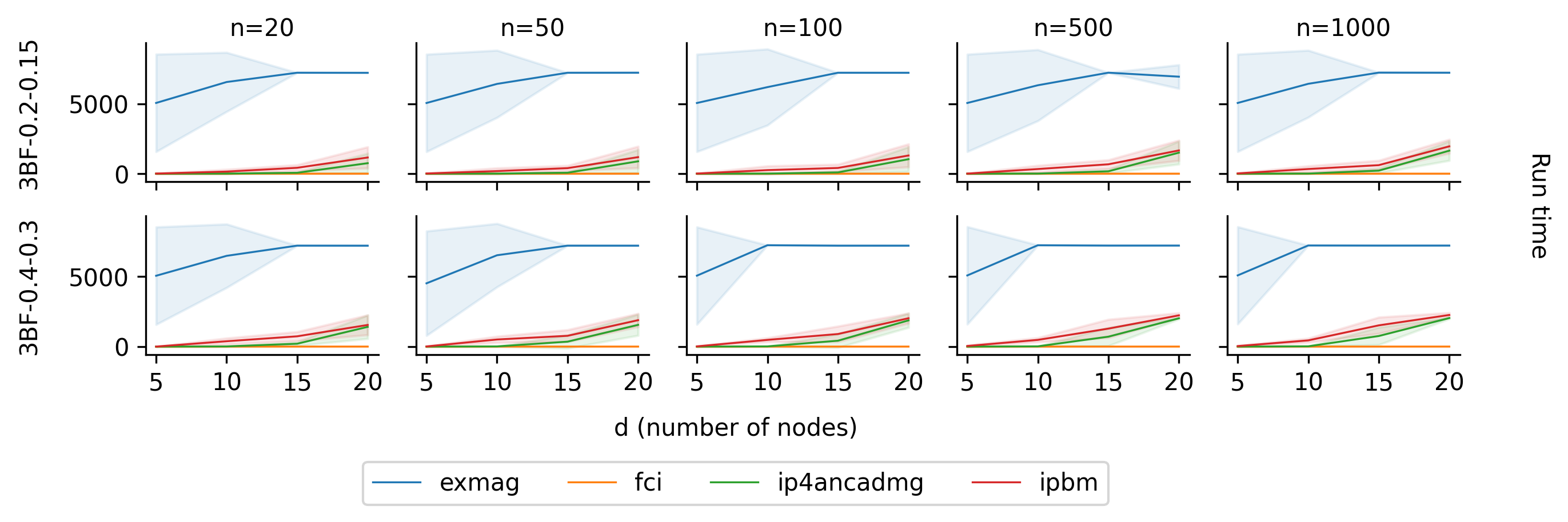}
    \caption{Run times of the compared algorithms in seconds.}
    \label{fig:runtime}
\end{figure*}

\begin{figure*}
    \centering
    \includegraphics[width=0.35\linewidth]{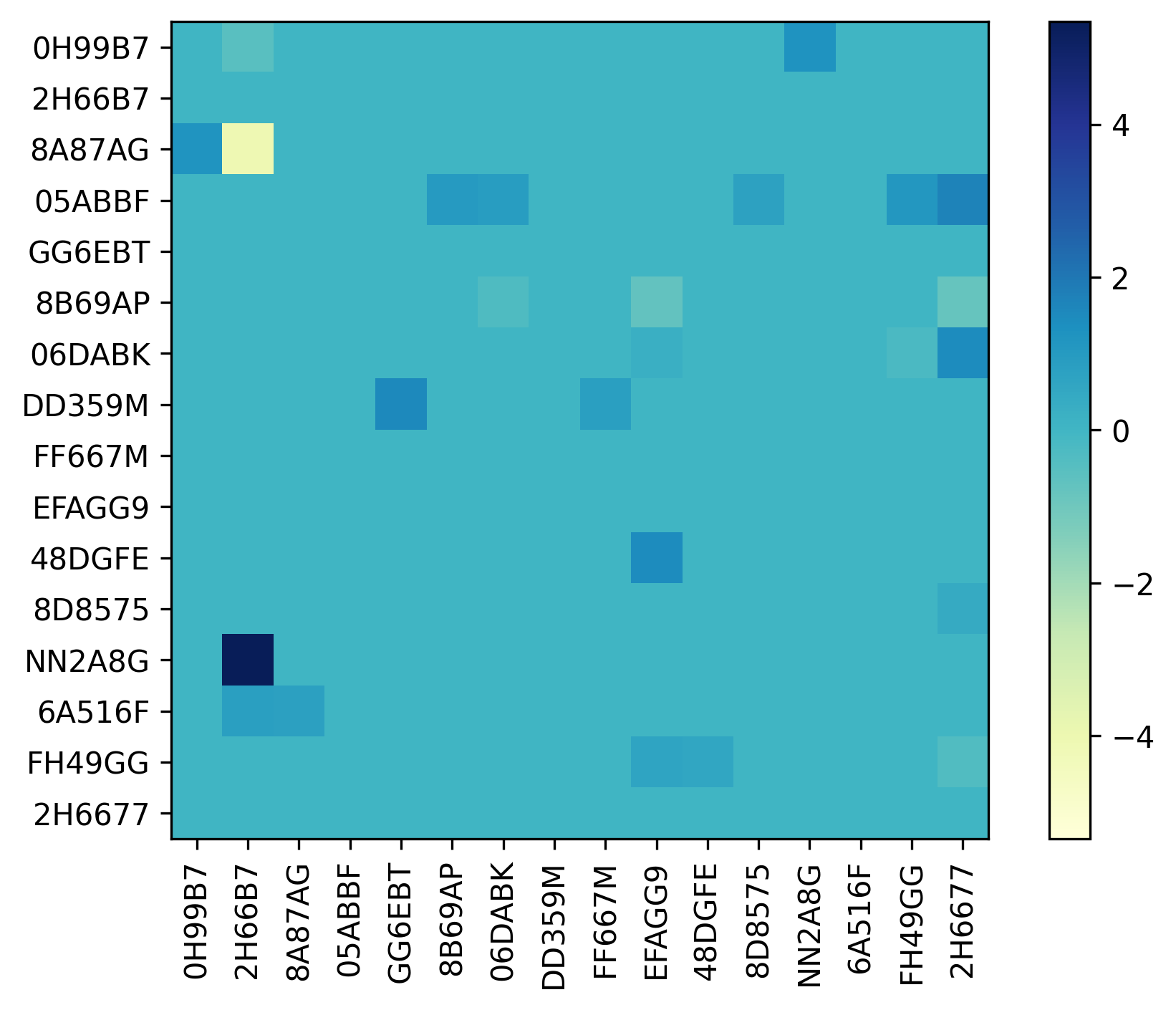}
    \includegraphics[width=0.35\linewidth]{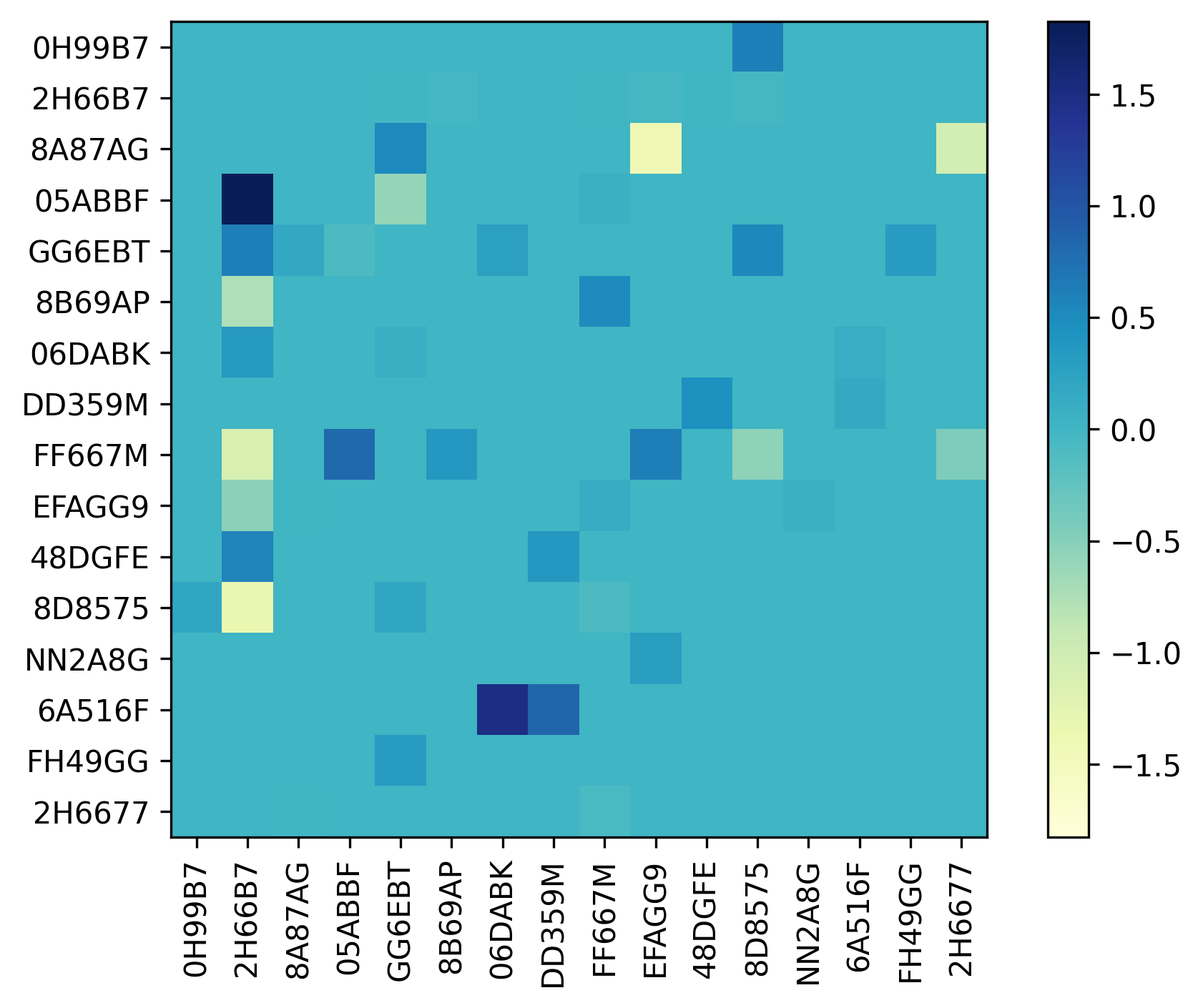}
    \caption{Heatmap of weight matrix $W$ (left) and bidirectional weight matrix $W$ (right) on the financial dataset.}
    \label{fig:real-world}
\end{figure*}

In the case of inducing paths, we use a DFS starting from each vertex. Once started from vertex $s$, the DFS routine checks for all possible inducing paths that terminate in $s$. For efficiency, a set of all possible endpoints of the path is held. Once this set is empty, the DFS search is terminated, and no further exploration is performed. The set is updated using the distance matrix calculated on the directed edges. If we consider a vertex $v$, $s$ must be either its ancestor (meaning that possible endpoints for $v$ remain unchanged) or the second inducing path endpoint is among the points that are reachable from $v$ (meaning that the possible endpoints for $v$ are replaced with their intersection with the set of all points reachable from $v$). See Algorithm \ref{alg:inducingpaths} for a simplified pseudocode that illustrates the idea, while leaving out some borderline cases and pruning conditions.

Once having the bidirected edges in the inducing paths and almost directed cycles, we need to trace back the Floyd-Warshall distance matrix to find all directed edges that form the cycle or the ancestor relationship. This is done using calls to the function \textsc{Trace-Distance-Matrix}, which can be found in the Supplementary materials.

If directed cycles, almost directed cycles, and inducing paths are found, the algorithm applies lazy constraints in \ref{constr:Acycliclazy} and in \ref{constr:InducingPathslazy}. Note that removing one directed edge between two vertices where multiple paths exist is not a necessary condition for the graph to become a MAG; however, this procedure can be repeated iteratively. If no inducing paths or almost directed cycles are found, we know that the program converged to the optimum, and we have a maximal ancestral graph, which minimizes \eqref{obj}.


\section{Experimental Evaluation}
\label{sec:experiments}

\paragraph{Used Datasets}

We tested the ExMAG algorithm on both synthetic and real-world datasets. The first synthetic data set is based on the \emph{Erd\"os-R\'enyi model} (ER) \cite{erdds1959random, zheng2018dags}, in which the ground truth graph is randomly selected from all graphs with $d$ vertices and $m$ edges (parameter of the experiment, for example, dataset ER-2 contains $2$ edges per variable, that is, $m = 2 \cdot d$). The weights of the graph are randomly sampled from the set $(-2.0, -0.5) \cup (0.5, 2.0)$.

Once the ground truth model is created, the training data are generated using the structural model equation. Then, $20\,\%$ of variables are treated as latent variables and hidden from the training data. The respective columns and rows from the ground truth weight matrix $W$ have also been removed. For fair comparison, $F$ matrix is not provided to ExMAG.

The second dataset uses randomly generated \emph{bow-free} (BF) graphs, based on paper \cite{bhattacharya2021differentiable}. A bow-free graph is a graph such that for no pair of vertices $i,j$,  $i \rightarrow j$ and at the same time $i \leftrightarrow j$.
The BF graph generation process has two parameters: the probability of a directed edge and the probability of a bidirected edge. The generation process is as follows. 
First, a bow-free graph with the given edge probabilities is generated randomly. Then the weights of the sampled graphs are randomly sampled from the set $(-2.0, -0.5) \cup (0.5, 2.0)$. As this work focuses in MAGs and not all BF graphs are MAGs, we filter out all BF graphs that are not MAGs from the test data.

The third synthetic dataset, 3BF, is a modified version of BF. We generate the ground truth graph in the same way as in the BF dataset and then modify it. Specifically, we identify each vertex with a degree greater than three, and randomly remove edges until the vertex has a degree of at most three. 

The adjacency matrix of the directed edges defines the weights of the structural equation model. 
Then the data samples are generated using the structural equation, where the noise is sampled from a multivariate Gaussian distribution with a covariance matrix equal to the adjacency matrix of bidirected edges generated in the previous step.

The fourth dataset uses real-world data from the \emph{financial} sector. \cite{BALLESTER2023101914} work with systemic credit risk, one of the most important concerns within the financial system, using dynamic Bayesian networks. The data show that transport and manufacturing companies are likely to transfer risk to other sectors, while banks and the energy sector are likely to be influenced by the risks from other sectors. The data from \cite{BALLESTER2023101914} contain a 10-time series capturing the spreads of 10 European credit default swaps (CDS), and further six time series are added from \cite{Rytivr2024exdag}.

We set the matrix $F$ to encode for no direct causal relationship between any two pairs of companies from different sectors. Banks sector includes 48DGFE, 05ABBF, 8B69AP, 06DABK, EFAGG9, 2H6677, FH49GG, and 8D8575. The insurance sector includes GG6EBT, DD359M, and FF667M. And lastly, the transportation sector and manufacturing include 0H99B7, 2H66B7, 8A87AG, NN2A8G, and 6A516F.

\paragraph{Evaluation Criteria}
\label{subsection:Criteria}
Suppose that a tested algorithm produced weight matrix $\hat{W}$. Such a matrix can contain nearly zero weights. For such reasons, thresholding is done, keeping only edges with a weight greater than or equal to $\delta$. In cases when the ground-truth weight matrix $W$ is known, the best solution (in terms of structural Hamming distance, see below) is kept over those defined by different threshold $\delta$ values.
In the evaluation, we use the \emph{structural Hamming distance (SHD)}. This distance is the sum of contributions over all pairs of variables in the graph. For two variables $i,j$, let $GT \in \{\rightarrow, \leftarrow, \leftrightarrow, \emptyset\}$ be the edge type in the ground truth graph and $PR \in \{\rightarrow, \leftarrow, \leftrightarrow, \emptyset\}$ be edge type in the predicted graph. Then the contribution of $i,j$ pair to SHD is $r_{ij} = 0$ if $GT = PR$, $0.5$ if $GT \neq PR \wedge GT \neq \emptyset \wedge PR \neq \emptyset$, and $1$ otherwise.
Other measured criteria include \emph{runtime} and \emph{F1-score}, i.e., the harmonic mean of precision and recall.

\paragraph{Experiment Setting}
\label{subsection:Setting}
In the experiments, we show the results of ExMAG. In the case of synthetic datasets, we generated random graphs with the number of vertices $d \in \{5, 10, 15, 20, 25\}$. The number of samples was $n \in \{20, 50, 100, 500, 1000\}$, and for the ground-truth graph, the edge-to-vertex ratio was in $\{2, 3, 4, 5, 6, 7, 8, 9, 10, 15, 20\}$. All tested algorithms were run $10$ times, each time on synthetic data generated using a random generator initialized with a different seed. The results were then averaged.
We compared our method with the FCI algorithm~\cite{spirtesCausalInferencePresence1995}, ip4ancadmg \cite{chen2021integer}, and ipbm \cite{dash2025integer}. We set regularization coefficient $\lambda$ to $1.0$. We ran experiments on a computing cluster with AMD EPYC 7543 cpus and each job had allocated two cores and 64GB RAM. Time limit was 900 seconds for ExMAG and 1800 seconds for other methods. The total cpu time needed for experiments in this paper was around one month.

\paragraph{Experimental Results}

The SHD results are shown in Figure \ref{fig:bstshd} and in the supplementary materials for additional datasets. The plots show a comparison of SHD values for ExMAG on the synthetic datasets. As can be seen, the structural Hamming distance grows with the number of variables.

In the case of ExMAG, a counterintuitive phenomenon happens on some of the datasets - the results get worse with the growing number of samples. This is caused by the interaction between the solver and the resulting MIQP - with the growing number of samples, the MIQP has a larger objective function in \eqref{obj}, resulting in slower evaluation, and thus it is more difficult for the optimizer to reach the global optimum. A simple countermeasure lies in downsampling the input data. To provide a fair comparison, Figure \ref{fig:bstshd} contains the results for the number of samples that gave the smallest SHD.

Since both ipbm and ip4ancadmg have a preprocessing step that depends exponentially on the maximum in-degree of the underlying ground truth graph, we tested these two algorithms only on the 3BF datasets, where the in-degree is bounded by three. Plot \ref{fig:bstshd} illustrates that to provide comparable results, ExMAG requires fewer samples than its competitors.
The run times of the evaluated algorithms are shown in Figure~\ref{fig:runtime}.
For additional results (incl. the F1-score), please see the supplementary materials.



The results on the real-world dataset can be seen in  Figure~\ref{fig:real-world}. Contrary to the original expectations, the highest risk importer is company 2H66B7, which stands for Lufthansa. The second highest risk importer is 2H6677, i.e., the Deutsche Bank, which is an expected result.

\section{Conclusion and Limitations}
\label{sec:conclusion}

Learning of Bayesian networks has received considerable attention as a means of causal learning. With a few exceptions, the research has not considered confounding explicitly. Our method, ExMAG, estimates a maximally ancestral graph, capturing confounding and causal relationships using bidirected and directed edges in a mixed graph. The method provides state-of-the-art statistical performance.

As with many other methods for causal learning, the scalability of the method may leave space for improvement. Although the branch-and-cut algorithm runs in time that is exponential in the number of variables in the worst case, Figure \ref{fig:runtime} illustrates that our algorithm is able to produce results comparable to SOTA in the time limit. Note that the reported runtime of ExMAG may overestimate the time required to obtain high-quality solutions, as the MIQP solver is allowed to use the full time budget to certify optimality. In practice, near-optimal solutions are often found much earlier, with additional time spent on proving optimality.
One could improve upon the run-time further by introducing additional cutting planes and more elaborate data structures for the separation of \ref{constr:Acycliclazy} and \ref{constr:InducingPathslazy}, perhaps drawing inspiration from solvers \cite{cook2011traveling} for the travelling salesman problem.

In terms of future work, exploring the predictive power of forecasting using variants of dynamical Bayesian networks with confounding considerations seems prominent. Although it seems clear that marginalization is hard even in dynamical Bayesian networks, and thus the computational complexity may be high, but  statistical performance is likely to improve, when confounding is considered.

\section*{Aknowledgements}
This work has received funding from the European Union’s
Horizon Europe research and innovation programme under grant
agreement No. 101084642. 
The work of G. K. has been partially supported by project MIS 5154714 of the National Recovery and Resilience Plan Greece 2.0 funded by the European Union under the NextGenerationEU Program.

\section*{Disclaimer}
This paper was prepared for information purposes and is not a product of HSBC Bank Plc. or its affiliates. Neither HSBC Bank Plc. nor any of its affiliates make any explicit or implied representation or warranty and none of them accept any liability in connection with this paper, including, but not limited to, the completeness, accuracy, reliability of information contained herein and the potential legal, compliance, tax or accounting effects thereof. Copyright HSBC Group 2025.

\clearpage 
\bibliography{refs}
\bibliographystyle{plain}

\clearpage

\appendix

\section{F1-score Results}

\begin{figure*}[h]
    \centering
    \includegraphics[width=0.7\linewidth]{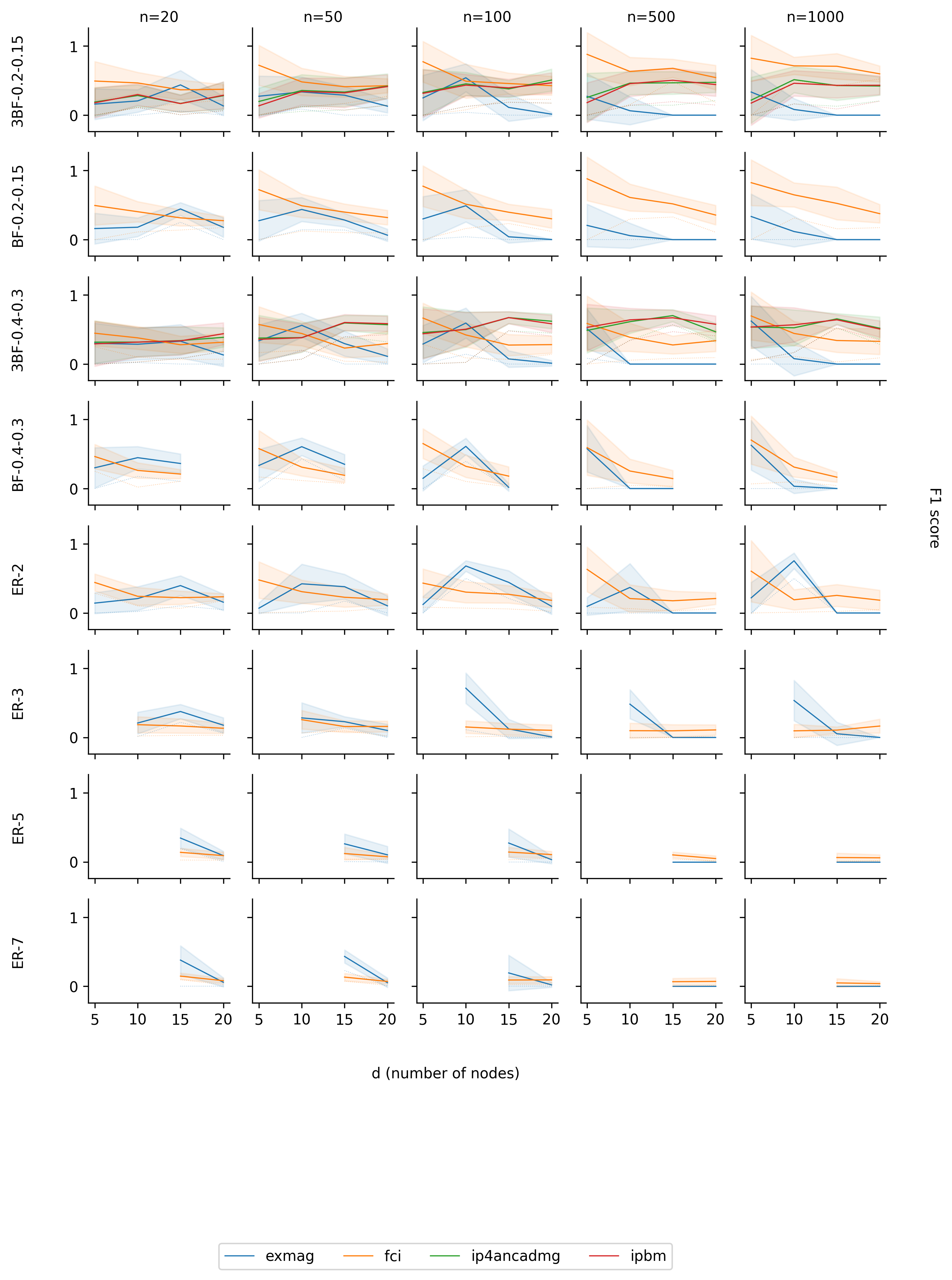}
    \caption{F1-score (in the vertical axis) for different settings of $d$ (in the horizontal axis) and $n$ (horizontal choice of the graph). The plots in the vertical dimension differ according to the dataset used.
    Standard deviations are depicted as the blurred regions, and dashed lines are the minimum values.
    Please note that for some of the ER plots, the graphs can be generated only for higher numbers of variables. For example, there exists no ER-5 with $d= 10$, as it would need to contain $50$ edges, while the maximum is $45$.
    }
    \label{fig:f1sm}
\end{figure*}


\clearpage
\section{SHD Results}


\begin{figure*}[h]
    \centering
    \includegraphics[width=0.7\linewidth]{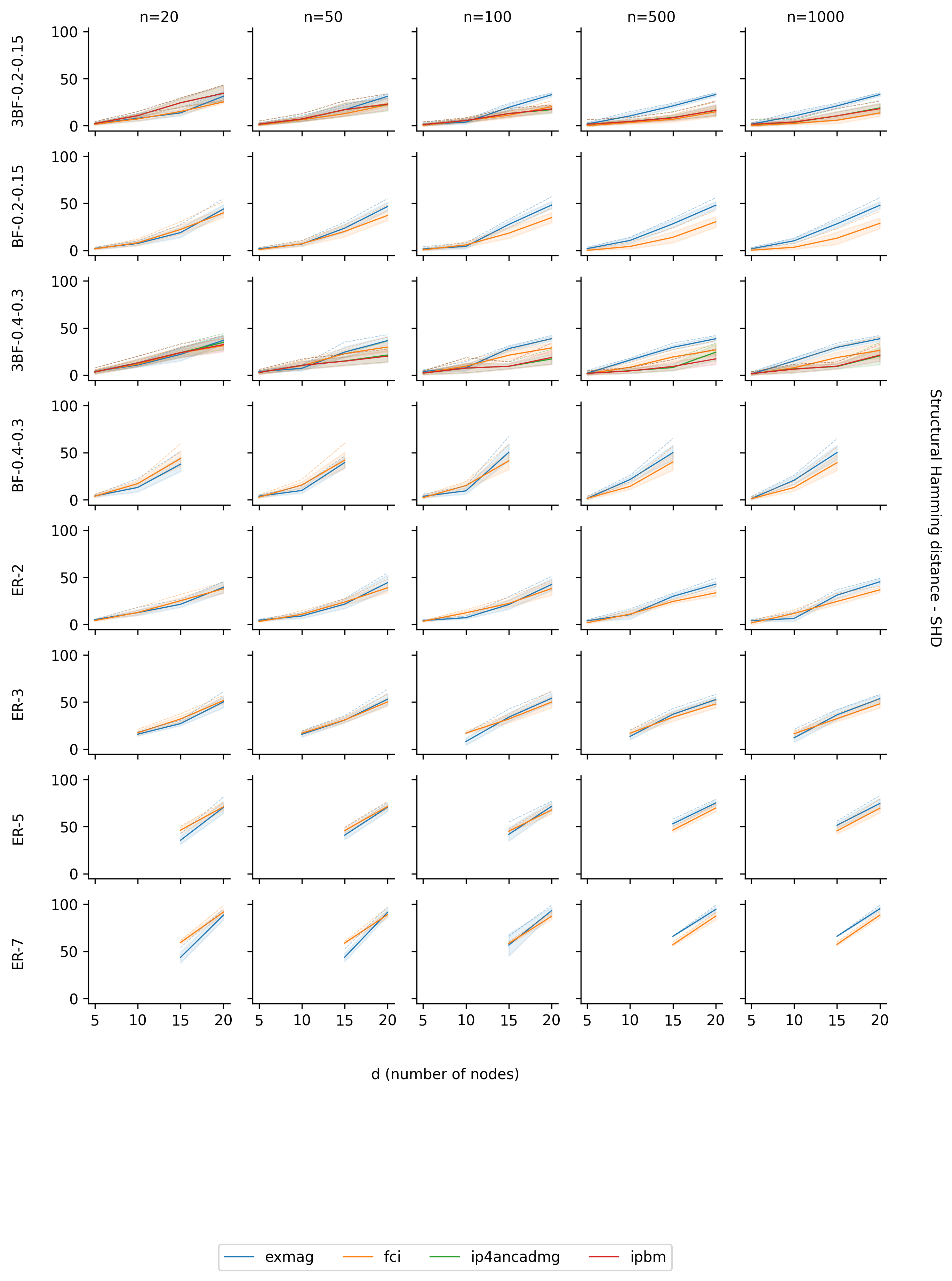}
    \caption{SHD values (in the vertical axis) for different settings of $d$ (in the horizontal axis) and $n$ (horizontal choice of the graph). The plots in the vertical dimension differ according to the dataset used. Standard deviations are depicted as the blured regions and dashed lines are the maximum values.
    }
    \label{fig:bstshdsm}
\end{figure*}

\clearpage
\section{Table of Notation}
\begin{table}[h]
\caption{A Table of Notation.}
\label{table:1}
\centering
\vskip 6pt
\begin{tabular}{| c | c | }\hline
    Symbol & Representation \\ 
    \hline
    $G$ & Gender \\
    $D$ & Department \\ 
    $A$ & Admit acceptance \\ 
    $U$ & potential confounder, is academic ability in this example \\ 
    \hline
    $\mathcal{G}$ & Graph \\
    $\mathcal{V}$ & Vertex set of the graph \\ 
    $\mathcal{E}$ & Set of directed edges \\ 
    $\mathcal{B}$ & Set of bidirected edges \\ 
    $\text{pa}_\mathcal{G}(v)$ & Parents of vertex $v$ in graph $\mathcal{G}$ \\ 
    $\text{sp}_\mathcal{G}(v)$ & Spouses of vertex $v$ in graph $\mathcal{G}$ \\ 
    $\text{an}_\mathcal{G}(v)$ & Ancestors of vertex $v$ in graph $\mathcal{G}$ \\
    \hline
    $W_D$ & Directed eight matrix \\ 
    $W_B$ & Bidirected eight matrix \\ 
    $E$ & Directed edge matrix \\ 
    $B$ & Bidirected edge matrix \\ 
    $F$ & Idicates that two variables have no direct causal relationship\\
    $X_{i,j}$ & Value of the $j$-th variable for the $i$-th data point \\ 
    ${w_D}_{k,j}$ & Weight of the directed edge from variable $k$ to variable $j$ \\ 
    ${w_B}_{k,j}$ & Weight of the bidirected edge from variable $k$ to variable $j$ \\ 
    $e_{j,k}$ & Binary variable indicating a directed edge from $j$ to $k$ \\ 
    $b_{j,k}$ & Binary variable indicating a bidirected edge between $j$ and $k$ \\ 
    $f_{j,k}$ & Binary variable indicating a existence of directed edge between $j$ and $k$ \\
    $d_{j,k}$ & Binary variable indicating a directed edge between $j$ and $k$ \\ 
    $r_{ij}$ & Contribution of pair $(i,j)$ to SHD \\ 
    $\lambda$ & Regularization parameter \\ 
    $c$ & Large constant for weight bounding \\ 
    $d$ & Number of variables \\ 
    $n$ & Number of data points \\ 
    $GT$ & Edge type in the ground truth graph \\ 
    $PR$ & Edge type in the predicted graph \\ 
    $\delta$ & Threshold for edge weights \\ 
    \hline
    $X$ & Causal variable \\ 
    $\epsilon$ & Noise term \\
    $\mathcal{P}$ & Class of distributions \\ 
    $C$ & Conditioning set \\ \hline

\end{tabular}
\end{table}

\section{Pseudocode}

\begin{algorithm}[H]
   \textbf{Input:} distances $D$ defined by directed edges $E$, start point $j$, and endpoint $k$ \\
   \textbf{Output:} edges on any path from $j$ to $k$
   \hrule
  \begin{algorithmic}
    \Function{Trace-Distance-Matrix}{$D$, $E$, $j$, $k$}
      \If{$D_{j,k} == \infty$} \Return ${}$
      \EndIf
      \State visited = $\{ (j, k) \}$
      \State stack = stack with $(j,k)$
      \State edges = $\{\}$
      \While{stack is not empty}
        \State $u, v \gets $ \Call{Pop}{stack} 
        \ForAll{$w \in {1,2, \ldots, d}$ s t. $D_{u,w}+D_{w,v} < \infty$}
          \State visited $\gets$ visited $\cup \{ (u, w), (w,v) \}$
          \If{$E_{u,w}$} edges $\gets$ edges $\cup \{ (u, w) \}$
          \ElsIf{$E_{w,v}$} edges $\gets$ edges $\cup \{ (w,v) \}$
          \EndIf
          \State add $\{ (u, w), (w,v) \}$ to stack
        \EndFor
      \EndWhile
      \Return edges
    \EndFunction
    \State
    \Function{Found-Inducing-Path}{$D$,$E$, $P$}
      \State $E'$ = $\{\}$
      \ForAll{vertices $j \in $ $P$ \mbox{ and } $j \not\in \{ P_0, P_{|P|}\}$}
        \State {\small $E' = E' \cup$ \Call{Trace-Distance-Matrix}{$D$,$E$,$j$,$P_0$} $\cup$ \Call{Trace-Distance-Matrix}{$D$,$E$,$j$,$P_{|P|}$}}
        \State \Comment{Finds all edges on any $j$ to $P_0$ ($P_{|P|}$) path}
      \EndFor
      \State {Found inducing path formed by path $P$ and directed edges $E'$}
    \EndFunction
  \end{algorithmic}
  \caption{Functions that help in the separation routine.}
  \label{algo:trace}
\end{algorithm}

\section{Derivation of the optimization criterion}

We base our cost function on work \cite{bhattacharya2021differentiable}. The paper assumes structural equation model
\begin{equation}
X \leftarrow X W_D + \varepsilon.
\end{equation}
where the noise term $\varepsilon$ contains correlations between variables, represented by covariance matrix $\beta = \mathbb{E}(\varepsilon \varepsilon^T)$. Matrix $\beta$ encodes the cofounding. With the goal to optimization of the ABIC criterion, \cite{bhattacharya2021differentiable} then minimizes (up to regularization) term
\begin{equation}
LS = \sum_{j=1}^d \left\| X_{\cdot, j} - X {W_D}_{\cdot, j} - Z^{(j)} \beta_{\cdot, j} \right\|^2,
\label{eq:lseqref}
\end{equation}
where $Z_{\cdot,j}^{(j)} = 0$ and for $k \neq j$
\begin{equation}
Z_{i,k}^{(j)} = \sum_{l=1, l\neq j}^d \varepsilon_{i,l} \beta^{-T}_{l,k}.
\label{eq:Zdef}
\end{equation}
The error term is calculated as
\begin{equation}
\varepsilon_{i,j} = X_{i,l} - X_{i, \cdot} {W_D}_{\cdot, j}.
\end{equation}
The notation is adjusted and the terms are rearranged to match the main paper body. For any matrix, for example $X$, $X_{i,j}$ denotes field in $X$ on $i$-th row, and $j$-th column, $X_{i,\cdot}$ denotes $i$-th row, and $X_{\cdot, j}$ denotes $j$-th column. Please note that \cite{bhattacharya2021differentiable} has a typo in Algorithm 2, where $X$ and ${W_D}_{\cdot, j}$ are reversed (resulting in undefined matrix product).

Here, we show the derivation of \eqref{obj} from the least squares error \eqref{eq:lseqref}. As a by-product, we show interpretation of matrix $W_B$. First, we expand the L2-norm in \eqref{eq:lseqref}, to obtain
\begin{equation}
  LS = \sum_{j=1}^d \sum_{i=1}^n \left\| X_{i, j} - X_{i, \cdot} {W_D}_{\cdot, j} - Z^{(j)}_{i,\cdot} \beta_{\cdot, j} \right\|^2.
\end{equation}
As both sums above are independent, we can swap their order. We expand matrix multiplication of $Z^{(j)}_{i,\cdot} \beta_{\cdot, j}$, keeping in mind, that $Z_{\cdot,j}^{(j)} = 0$.
\begin{equation}
  LS = \sum_{i=1}^n \sum_{j=1}^d \left\|
  X_{i, j} - X_{i, \cdot} {W_D}_{\cdot, j} - \sum_{k=1, k\neq j}^d Z^{(j)}_{i,k} \beta_{k, j}
  \right\|^2.
\end{equation}
Further, we plug \eqref{eq:Zdef} into the formula
\begin{equation}
  LS = \sum_{i=1}^n \sum_{j=1}^d \left\|
  X_{i, j} - X_{i, \cdot} {W_D}_{\cdot, j} - \sum_{k=1, k\neq j}^d \left[ \sum_{l=1, l\neq j}^d \varepsilon_{i,l} \beta^{-T}_{l,k}  \right]
 \beta_{k, j}
  \right\|^2.
\end{equation}
As indices $k$ and $l$ are independent, we can rearrange the terms into
\begin{equation}
  LS = \sum_{i=1}^n \sum_{j=1}^d \left\|
  X_{i, j} - X_{i, \cdot} {W_D}_{\cdot, j} - \sum_{k=1, k\neq j}^d \sum_{l=1, l\neq j}^d 
  \left[ \varepsilon_{i,l} \beta^{-T}_{l,k}  \beta_{k, j}  \right]
  \right\|^2.
\end{equation}
Further, we can swap the order of summation, and factor out $\varepsilon_{i,l}$ to obtain
\begin{equation}
  LS = \sum_{i=1}^n \sum_{j=1}^d \left\|
  X_{i, j} - X_{i, \cdot} {W_D}_{\cdot, j} - \sum_{l=1, l\neq j}^d  \varepsilon_{i,l}
  \left[ \sum_{k=1, k\neq j}^d \beta^{-T}_{l,k}  \beta_{k, j}  \right]
  \right\|^2.
\end{equation}
Let us denote $W_B$ matrix obtained by the multiplication of matrices $\beta^{-T}\beta$ which uses the summation above, i.e., let
\begin{equation}
  {W_B}_{l,j} = \sum_{k=1, k\neq j}^d \beta^{-T}_{l,k}  \beta_{k, j}.
  \label{eq:wbdef}
\end{equation}
Then, using notation from \eqref{eq:errorR}, we obtain
\begin{equation}
  LS = \sum_{i=1}^n \sum_{j=1}^d \left\|
  R_{i,} - \sum_{l=1, l\neq j}^d  R_{i,l}{W_B}_{l,j}
  \right\|^2,
\end{equation}
which is equivalent to the MIQP formulation in \eqref{obj}.

Now, we provide interpretation of the weight matrix $W_B$. Denote $\mathop{\mathrm{diag}}(\beta)$ the matrix containing containing diagonal entries of $\beta$, i.e., the diagonal matrix of marginal variances. As summation in \eqref{eq:wbdef} ignores fields, where $k=j$, i.e., we can replace $\beta_{k, j}$ with $0$ for $k=j$. As a result $W_B$ can be written in matrix form as
\begin{equation}
W_B = \beta^{-T}(\beta - \mathop{\mathrm{diag}}(\beta)).
\end{equation}
This can be rearranged into ($I$ stands for identity matrix)
\begin{equation}
W_B = I - \beta^{-T}\mathop{\mathrm{diag}}(\beta).
\end{equation}
For non-diagonal entry of $W_B$, its value is zero only if the respective field in the precision matrix is zero. In other words, ${W_B}_{l,j} = 0$ if and only if $l$ and $j$ are conditionally independent given all variables but $l$ and $j$.

\section{Residual Assumptions And Statistical Significance}
There exists a fundamental premise in structural equation modeling that the residuals—representing unexplained variation—are asymptotically unbiased, meaning they are independent of both observed and latent variables, and follow a zero-mean distribution. This assumption plays a critical role in ensuring that learned causal relationships are not distorted by hidden confounders or systematic error. The ExMAG framework embraces this principle by design, explicitly modeling residual independence as a safeguard against spurious causal edges. Just as fairness-aware systems aim to isolate structural patterns from social bias \cite{barocas2019fairness}, ExMAG works to separate signal from statistical noise. The result is a model capable of learning causal mechanisms that are not only mathematically sound but also resilient across different subpopulations, forming a foundation of causal inference.

Causal discovery systems, like decision-making algorithms in high-stakes domains, must operate effectively across structurally diverse populations. This paper uses real-world financial data spanning multiple sectors—banking, insurance, manufacturing, and transportation—each exhibiting distinct systemic exposures. These domains can be viewed as a \textit{privileged setting} where data availability and quality are high, yet subgroup heterogeneity remains significant. In such contexts, ExMAG successfully identifies dominant risk propagation patterns, even when feature distributions vary across industries. This mirrors broader challenges in fairness: the need to perform robustly across populations with unequal baseline conditions \cite{mehrabi2021survey}. The model’s consistent recovery of risk links—illustrated in Figure~\ref{fig:real-world}—not only affirms its structural fidelity but also its capacity to generalize without group-specific tuning.

Understanding the statistical reliability of a model’s output requires more than average performance—it demands insight into variance. To that end, the authors conduct 10 independent trials for each configuration, reporting both mean and standard deviation for key metrics such as SHD and F1-score. The inclusion of error bars in Figures~\ref{fig:bstshd} and \ref{fig:f1sm} provides a visual representation of variability, revealing not just how well the model performs, but how consistently. In contrast to baseline methods with large fluctuations, ExMAG demonstrates narrow error margins, underscoring its stability in the face of stochastic elements like data partitioning and initialization.

\section{Brief Introduction To Mixed Integer Quadratic Programming}
Let us also provide a short introduction to mixed-integer quadratic programming. An optimization problem is called a mixed-integer quadratically constrained quadratic program (MIQCQP) if it is of the form
\begin{align}
\underset{x\in\mathbb{R}^{n}}{\min} \quad & x^{T}Qx+q^{T}x,\label{eq:qp_cost_function} \\
 \text{s.t. }&x^{T}Q_{i}x+q_{i}^{T}x\leq a_{i}, \label{eq:qp_quadratic_const} \\
 &Ax \leq b, \label{eq:qp_linear_const} \\
 &x \in \mathbb{R}^{n-r}\times \mathbb{Z}^{r}\label{eq:qp_integrality}
\end{align}
where $Q, Q_{i} \in \mathbb{R}^{n\times n}$, $q, q_{i} \in \mathbb{R}^n$, $A \in \mathbb{R}^{m\times n}$, $a \in \mathbb{R}^k$, $b \in \mathbb{R}^m$
and $m,n,k,r \in \mathbb{N}$. \eqref{eq:qp_cost_function} is called the cost or loss function, \eqref{eq:qp_quadratic_const} represents the quadratic constraints, \eqref{eq:qp_linear_const} are the linear constraints, and \eqref{eq:qp_integrality} enforces the integrality constraints for the last $r$ components of the vector of decision variables $x$.

Mixed-integer quadratic programs have been shown to be NP-hard \cite{pia14}, which often leads to an exhaustive demand for computational resources. The algorithms used to solve MIQP are typically branch-and-bound or cutting plane \cite{Dakin1965ATA, bonami09, WESTERLUND1995131, kron15}. Both of these algorithmic treatments are often employed together, often with the addition of a presolving step, the use of heuristics, and parallelism. The aforementioned allows many modern solvers to solve even large problems despite the NP-hardness. Some of these solvers are open source (like SCIP and GLPK), and others are commercial (GUROBI and CPLEX). The powerful infrastructure present in these solvers can be made use of together with additional problem-specific modifications to deliver high-quality solutions.

Due to the exhaustive nature of the algorithms mentioned in the previous paragraph, global convergence is guaranteed \cite{Belotti_Kirches_Leyffer_Linderoth_Luedtke_Mahajan_2013}. Furthermore, convergence to the global solution may be tracked and the error estimated by computing the dual problem of (\ref{eq:qp_cost_function}--\ref{eq:qp_integrality}). The dual of the problem is then used to compute the so-called MIP GAP as follows
\begin{equation}
\text{MIP GAP}=\frac{\left|J\left(x^{*}\right)-J_{\text{dual}}\left(y^{*}\right)\right|}{\left|J\left(x^{*}\right)\right|},    
\end{equation}
where $x^{*}$ and $y^{*}$ are the current best solutions of the primal and dual problems, respectively, and $J$ and $J^{*}$ are the cost functions of the primal and dual problems, respectively. The MIP GAP ensures that we can assess the quality of the minimization during solution time and terminate the computation when the result is good enough (small enough MIP GAP). Furthermore, if the gap reaches 0 at any point, we are sure that the current solution is a global optimum.

\end{document}